# Genetic Random Weight Change Algorithm for the Learning of Multilayer Neural Networks

Sarker Mohammad Ibrahim
Division of Electronics
Chonbuk National University
Jeonju, South Korea
sarkeribrahim@gmail.com

Yali Nie
Electronic Engineering
Chonbuk National University
Jeonju, South Korea
yali_nie@hotmail.com

Hong Youngki
National Institute of Agricultural Sciences
Jeonju, South Korea
sanm70@korea.kr

Hyongsuk Kim
Division of Electronics
Chonbuk National University
Jeonju, South Korea
hskim@jbnu.ac.kr

*Abstract*—A new method to improve the performance of Random weight change (RWC) algorithm based on a simple genetic algorithm, namely, Genetic random weight change (GRWC) is proposed. It is to find the optimal values of global minima via learning. In contrast to Random Weight Change (RWC), GRWC contains an effective optimization procedure which are good at exploring a large and complex space in an intellectual strategies influenced by the GA/RWC synergy. By implementing our simple GA in RWC we achieve an astounding accuracy of finding global minima.

## I. INTRODUCTION

Implementing neural network in hardware has long been a problem for the scientists and researchers. Although researchers have been engaged in fabrication of neural network hardware, only a few neural networks implemented with a learning algorithm have been reported [1]. The complexity of the algorithms is the main problem to implement a learning algorithm. Some researchers have fabricated on-chip learning neural networks. Arima et al. implemented a revised Boltzmann Machine learning algorithm with analog/digital hybrid circuitry [3], Morie et al. fabricated an analog neural network chip with Back Propagation algorithm [4]. Yasunaga et al. and Shima et al. [5],[6] employed digital circuitry for implementation of Back Propagation or Hebbian learning. These traditional learning algorithms are expressed with complex equations and require complicated multiplication. Unlike these all traditional algorithm, random weight change algorithm is easy and suitable for analog implementation even though random weight change algorithm is less efficient than back propagation and do suffer due to local minimum problems.

Genetic algorithm are inspired by Darwin's theory about evolution-"survival of the fittest" so in genetic algorithm it runs simultaneously in one generation and the algorithm always pick the best results in one generation and then produce the next generation[9]. Wang et al. [10] used genetic algorithm (GA) with artificial neural network (ANN) to find out optimal process parameters for optimal performances.

So, random weight change and genetic algorithm are the two techniques for optimization and learning, each with its own weaknesses, adding GA with RWC can make RWC more robust and can avoid local minima problems.

Combining different algorithms can fetch better results than they could achieve individually. In this paper, we propose a modification of random weight change (RWC) algorithm which allows us to combine genetic algorithm (GA) with random weight change (RWC) algorithm, we called it as genetic random weight change algorithm (GRWC). And the results obtained through this method shows an improvement than the conventional random weight change method.

The paper is organized as follows: section II & III gives an overview of random weight change (RWC) and genetic algorithm (GA). In Section IV, we proposed our genetic random weight change (GRWC) algorithm. In Section V & VI, the simulations results and conclusion are given.

## II. RANDOM WEIGHT CHANGE ALGORITHM

In RWC algorithm, the synaptic weights in the networks are going to be updated randomly from the initial states with a small increment of $\pm\lambda$. Due to weight change, if the summation of errors at the output decreases from the desired values than the same weight change is iterated until the error increases. However, if the summation of errors increases than again the weights are updated randomly. The weights updating algorithm in RWC is given below,

$$\Theta_i^{(1)}(n+1) = \Theta_i^{(1)}(n) + \Delta\Theta_i^{(1)}(n+1)$$

where

$$\Delta\Theta_i^{(1)}(n+1) = \Delta\Theta_i^{(1)}(n) \quad \text{if } J_i(n+1) < J_i(n)$$

$$\Delta\Theta_i^{(1)}(n+1) = \lambda \cdot rand \quad \text{if } J_i(n+1) \geq J_i(n)$$

When J(n) defines the cost function of the output, $\lambda$ is the small constant value of either +1 or -1. After performing the statistical descent the network finally reaches the global minima. This learning algorithm is less efficient than back propagation even though it is good for analog implementation [2], [1], [7].

### III. GENETIC ALGORITHM

Genetic algorithms (GA) are computer programs that mimic the processes of biological evolution in order to solve problems and to model evolutionary systems [8]. The simplest GA involves three main type of operators:

- *Selection:* This operator selects chromosomes in the population for reproduction. As fitter is the chromosome, it has more chances of being selected to reproduce.

- *Crossover:* During crossover GA creates new individuals by combining aspects of selected individuals. This operator creates two offspring of two chromosomes by exchanging sub sequences. For example, the values of the two strings
  000000 and 111111
  Could be crossed over after the third locus in each to produce the two offspring
  110000 and 001111.
  This two new offspring created from this mating are put into the next generation of the population. Recombining portions of good individuals, this process is likely to create even better individuals.

- *Mutation:* Mutation and selection (without crossover) create a parallel, hill-climbing algorithm. This operator randomly flips some bits in a chromosome. For example, the string 00000100 might be mutated in its second position to yield 01000100. Mutation can occur at each bit position in a string with some low probability, usually very small.

In most applications of GA the basis is the simple procedure described in this above section.      GA that works on representation other bit strings or GA that have different typed of crossover and mutation operators, are more complicated examples of GA. John Holland was the creator of the field of genetic algorithm [Holland 1975] was an early landmark and the best introduction for the interested reader is [Goldberg 1988].

### IV. GENETIC RANDOM WEIGHT change ALGORITHM

To design our proposed genetic random weight change algorithm, we took selection and mutations operators from genetic algorithm, though here mutation operator works on random weight change algorithm. In nature world, we cannot say an individual is strong or not when it is newly born. We should let it develop for a period of time and make the decision whether to let it breed or eliminate it. Therefore, in our algorithm, all the eight simulations run for 1000 epochs, then choose the best two and let them copy and reproduce. Our Genetic random weight change algorithm operates according to following steps

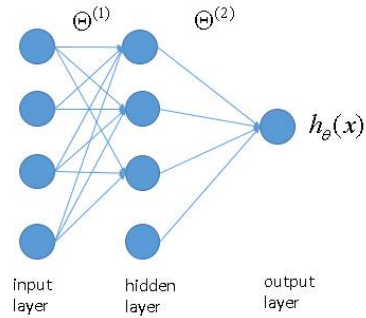

Fig. 1: - Structure of a neural network.

Symbols:

$x$ : *input layer value*

$\Theta$ : *weights*

$z_2$ : *hidden layer value*

$z_3$ : *output layer node value*

$h_\Theta$ : *prediction of label possibility on output layer*

$y$ : *real output label*

$f$ : *activation function on neural nodes* $f(x) = \dfrac{1}{1+e^{-x}}$

$\lambda$ : *limitation of weight increasement*

$\alpha$ : *learning rate of weight*

$N$ : *number of candidate in the genetic algorithm*

$N\_label$ : *number of labels*

$rand$ : *ramdom value from -1 to 1*

Steps of algorithm:

1. *Initialization:* A number of candidate weights of the neural network is generated, very often these have random values

$$\Theta_i^{(1)} = \lambda \cdot rand, \quad \Theta_i^{(2)} = \lambda \cdot rand, \quad i = 1, 2, \cdots, N$$

$$\Delta\Theta_i^{(1)} = \lambda \cdot rand, \quad \Delta\Theta_i^{(2)} = \lambda \cdot rand, \quad i = 1, 2, \cdots, N$$

2. *Evaluation*: Calculate cost function J for each candidate. A cost function will allow to score the prediction performance of each candidate; the score will be a number that tells how good this solution solves the problem.

$$z_{2,i} = f(\Theta_i^{(1)} \cdot x) \quad i = 1, 2, \cdots, N$$

$$z_{3,i} = f(\Theta_2^{(2)} \cdot z_{2,i})$$

$$h_{\Theta,i}^j = \frac{z_{3,i}^j}{\sum_{j=1}^{N\_label} z_{3,i}^j}$$

$$J_i = \frac{1}{2}(h_{\Theta,i} - y)^2$$

Select two best candidates from the groups according to the cost value each individual can achieve.

$$Index_{1,2} = \min(J)$$

3. Update neural network weight

    3.1 *Copy and Reproduce:* Copy the neural weight from the best 2 selected candidates. Abandon the value of other candidates and assign their weights from the best 2 candidates.

    $$\Theta_i^{(1)} = \Theta_{Index_1}^{(1)}, \quad i=1,2,\cdots,n/2$$
    $$\Theta_i^{(1)} = \Theta_{Index_2}^{(1)}, \quad i=n/2+1, n/2+2,\cdots,n$$
    $$\Theta_i^{(2)} = \Theta_{Index_1}^{(2)}, \quad i=1,2,\cdots,n/2$$
    $$\Theta_i^{(2)} = \Theta_{Index_2}^{(2)}, \quad i=n/2+1, n/2+2,\cdots,n$$

    $$\Delta\Theta_i^{(1)} = \Delta\Theta_{Index_1}^{(1)}, \quad i=1,2,\cdots,n/2$$
    $$\Delta\Theta_i^{(1)} = \Delta\Theta_{Index_2}^{(1)}, \quad i=n/2+1, n/2+2,\cdots,n$$
    $$\Delta\Theta_i^{(2)} = \Delta\Theta_{Index_1}^{(2)}, \quad i=1,2,\cdots,n/2$$
    $$\Delta\Theta_i^{(2)} = \Delta\Theta_{Index_2}^{(2)}, \quad i=n/2+1, n/2+2,\cdots,n$$

    3.2 *Random Weight Change:* With the mutation based on random weight change algorithm of all the candidates, we get the offspring weight and the cost value of each individual.

    $$\Theta_i^{(1)}(n+1) = \Theta_i^{(1)}(n) + \Delta\Theta_i^{(1)}(n+1)$$
    $$\Theta_i^{(2)}(n+1) = \Theta_i^{(2)}(n) + \Delta\Theta_i^{(2)}(n+1)$$

where

$$\Delta\Theta_i^{(1)}(n+1) = \Delta\Theta_i^{(1)}(n) \quad \text{if } J_i(n+1) < J_i(n)$$
$$\Delta\Theta_i^{(1)}(n+1) = \lambda \cdot rand \quad \text{if } J_i(n+1) \geq J_i(n)$$

    *3.3 Test:* Test the cost function value and prediction correction based on test data.

4. By repeating following steps until the cost goes down below satisfactory value.

The general idea of combining GA and RWC is illustrated in this figure 2.

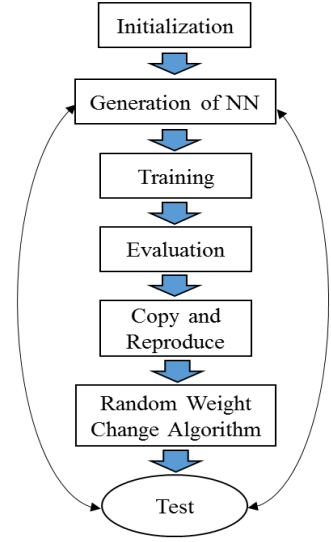

Fig. 2: -The Principle Structure of GRWC Algorithm.

## V. RESULTS

To verify the learning performance of our GRWC algorithm, we trained it with 100 images of MNIST dataset. Also, we compare it with existing RWC algorithm for handwritten character recognition problem. Both RWC and GRWC neural architecture contains a hidden layer with five hidden nodes and ten output nodes. As we mentioned earlier that GRWC performed eight simulations parallel for 1000 epochs than choose the best two simulated results, and let them copy and reproduce the process. For this reason, the total average iterations should be $1.32*10^5 *8=1056*10^5$. However, in RWC case the total average iterations $4.309*10^6$. Figure 3 illustrate the average error curves for both RWC and GRWC algorithm. It is distinctly observable that the average error per iterations for GRWC is less than RWC algorithm. Moreover, for better illustration we add the table I to compare the average error and iterations of GRWC and RWC.

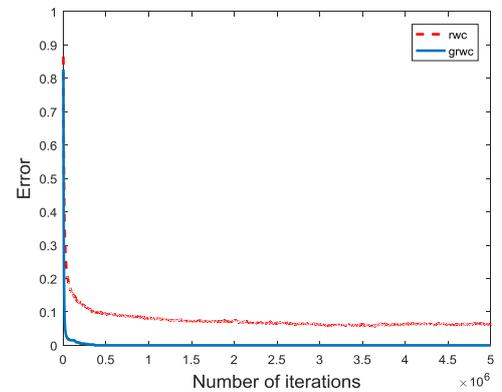

Fig. 3: - Average error curve for RWC & GRWC.

TABLE I.

| Simulation number | RWC iterations | RWC error | GRWC iterations | GRWC error |
|---|---|---|---|---|
| 1 | $5*10^6$ | 0.0379 | $1.8*10^5$ | 0.01 |
| 2 | $5*10^6$ | 0.1725 | $5.1*10^4$ | 0.01 |
| 3 | $5*10^6$ | 0.0589 | $3.6*10^4$ | 0.01 |
| 4 | $5*10^6$ | 0.0448 | $1.97*10^5$ | 0.01 |
| 5 | $5*10^6$ | 0.1510 | $3.56*10^5$ | 0.01 |
| 6 | $5*10^6$ | 0.0764 | $2.57*10^5$ | 0.01 |
| 7 | $5*10^6$ | 0.0331 | $5.4*10^4$ | 0.01 |
| 8 | $3.063*10^6$ | 0.0196 | $1.43*10^5$ | 0.01 |
| 9 | $5*10^6$ | 0.0450 | $3.4*10^4$ | 0.01 |
| 10 | $2.88*10^4$ | 0.0193 | $1.9*10^4$ | 0.01 |
| average | $4.309*10^6$ | 0.0498 | $1.32*10^5$ | 0.01 |

By analyzing the above table I, it was found that GRWC shows far better learning curve than RWC. A total of 10 simulation cases were performed by both methods. In case of GRWC it reached the global optima (0.01) every time, whereas RWC reached to global optima only twice. As there was no definite trend for number of iterations, an average value of iteration and error for both methods were calculated. The average value of iteration for GRWC was lower than RWC.

## VI. CONCLUSION

In this study we presented the GRWC algorithm which can be effectively used against RWC to reach to the global optima. The simulation results showed that it is possible to use genetic algorithms in RWC to avoid local optima. The average error curve illustrates that the learning in GRWC is more accurate than the RWC as the error reached to global optima (0.01). The efficiency of GRWC is far better than RWC as it reaches the minimal value each and every time where's the RWC efficiency is only 20% over a 10 simulation results. Also considering the average value of iterations, it was seen that GRWC takes less iteration than RWC to reach global optima. So GRWC algorithm is more accurate and efficient than RWC and can be a better option to use in circuits. Hence GRWC being a hybrid algorithm adds up one more option for users working in ANN.

Acknowledgment

This work was carried out with the support of "Cooperative Research Program for Agriculture Science and Technology Development (Project No. PJ1603000500)" Rural Development Administration, Republic of Korea and BK 21 Plus, Chonbuk National University.

References

[1] Kenichi Hirotsu, Martin A. Brooke, An Analog Neural Network Chip with Random Weight Change Learning Algorithm, Proceedings of 1993 International Joint Conference on Neural Networks.

[2] Shyam Prasad Adhikari1, Hyongsuk Kim2, Ram Kaji Budhathoki2, Changju Yang2 and Jung-Mu Kim2, Learning with Memristor Bridge Synapse-Based Neural Networks.

[3] Y. Arima et al., " A self-learning Neural Network Chip with 125 Neurons and 10K Self-organization Synapses", IEEE Journal of Slid-State Circuits, vol. 26, no. 4, pp. 607, 199

[4] T. Morie et al., " Analog VLSI Implementation of adaptive algorithms by an extended H synapse circuit", IEICE Transactions on Electronics, vol. €575, no. 3, pp. 303, 1992

[5] T. Shima et al., " Neuro Chips with On-Chip Back-Propagation and/or Hebbian Learning", IEEE Journal of Solid-state Circuits, vol. 27, no. 12, 1992.

[6] M. Yasunaga et al., " A Self-Learning Digital Neural Network Using Wafer-Scale LSI", IEEE Journal of Solid-state Circuits, vol. 28, no. 2, pp. 106, 1993.

[7] B. Burton, F. Kamran, R. G. Harley and T. G. Habetler, M. A. Brooke and R. Poddar, ''Identification and control of induction motor stator currents using fast online random training of neural network,'' IEEE Trans. Ind. Applicat., vol. 33, May 1997.

[8] Melanie Mitchell ,Genetic Algorithms: An Overview, Santa Fe Institute, Complexity, 1 (1) 31–39, 1995

[9] Grefenstette, J. 1986. Optimization of control parameters for genetic algorithms. IEEE transactions on Systems, Man, and Cybernetics 16(1):122-128.

[10] K. Wang, H.L. Gelgele, Y. Wang, Q. Yuan, M. Fang, A hybrid intelligent method for modelling the EDM process, Int. J. Machine Tools Manuf. 43 (2003) 995–999.